# AC-MAMBASEG: AN ADAPTIVE CONVOLUTION AND MAMBA-BASED ARCHITECTURE FOR ENHANCED SKIN LESION SEGMENTATION

**Viet-Thanh Nguyen, Van-Truong Pham\*, Thi-Thao Tran**

Department of Automation Engineering, School of Electrical and Electronic Engineering
Hanoi University of Science and Technology, Hanoi, Vietnam
thanh.nv202524@sis.hust.edu.vn; thao.tranthi@hust.edu.vn; truong.phamvan@hust.edu.vn

## Abstract

Skin lesion segmentation is a critical task in computer-aided diagnosis systems for dermatological diseases. Accurate segmentation of skin lesions from medical images is essential for early detection, diagnosis, and treatment planning. In this paper, we propose a new model for skin lesion segmentation namely AC-MambaSeg, an enhanced model that has the hybrid CNN-Mamba backbone, and integrates advanced components such as Convolutional Block Attention Module (CBAM), Attention Gate, and Selective Kernel Bottleneck. AC-MambaSeg leverages the Vision Mamba framework for efficient feature extraction, while CBAM and Selective Kernel Bottleneck enhance its ability to focus on informative regions and suppress background noise. We evaluate the performance of AC-MambaSeg on diverse datasets of skin lesion images including ISIC-2018 and PH2; then compare it against existing segmentation methods. Our model shows promising potential for improving computer-aided diagnosis systems and facilitating early detection and treatment of dermatological diseases. Our source code will be made available at: *https://github.com/vietthanh2710/AC-MambaSeg*.

***Keywords***: Skin Lesion Segmentation. Visual Mamba. Convolution. CBAM. Attention gate. Selective Kernels.

## 1. INTRODUCTION

Dermatological diseases, including skin cancer, present significant challenges in medical diagnosis and treatment due to their varied manifestations and potential severity. According to [1], in 2019 alone, the global mortality from skin and subcutaneous diseases amounted to 98,522 (95% uncertainty interval 75,116 – 123,949) individuals. The burden of these diseases, as measured by disability-adjusted life years (DALYs), was estimated at 42,883,695. Of this burden, 5.26% was attributed to years of life lost (YLLs), while the remaining 94.74% was due to years lived with disability (YLDs). Therefore, the improvement of tasks related to dermatological image processing is always a matter of great concern, among which the segmentation task cannot be overlooked.

Accurate segmentation of skin lesions images is essential for computer-aided diagnosis systems, as it enables precise quantification of lesion characteristics and facilitates the development of automated diagnostic tools. Compared with MRI or CT images acquired by advanced technology that are more precise and clearer, the opposite is true for skin lesion images. Skin lesion segmentation is a challenging task due to the inherent variability in lesion appearance, size, texture, and context. Traditional segmentation methods often struggle to handle the diverse and intricate features of skin lesions, leading to suboptimal results and limited clinical utility.

Initially, approaches to skin lesion segmentation often rely on conventional convolutional neural networks (CNNs) to capture relevant features and improve segmentation accuracy [2]–[5]. So far, segmentation models using only CNNs have begun to reveal weaknesses because of a lack of global feature capture ability. Researchers have carried about attention and then brought Transformers [6] from natural language processing (NLP) field into computer vision via Vision Transformer (ViT) [7], causing an explosional trend when they are able to capture long-range dependencies in images and focus on important image regions, enabling them to consider global context information and facilitate more effective feature extraction.

In the realm of medical image segmentation, TransUNet [8] emerged as a pioneering approach by integrating Transformers alongside CNNs within the encoder of a U-shaped architecture. Demonstrating its efficacy, Gulzar and Khan revealed that TransUNet surpassed several models based solely on CNNs for skin lesion segmentation [9]. Addressing computational challenges associated with high-resolution medical images, Cao et al. introduced the Swin-Unet [10], which incorporates self-attention mechanisms within shifted windows. Furthermore, there were a number of notable skin segmentation studies utilizing Transformers, such as BAT [11], FAT-Net [12]. Notwithstanding the above advantages, Transformers require expensive computational resources when dealing with large-scale datasets or high-resolution images, resulting in a significant processing burden, especially for tasks like medical image segmentation that require dense predictions. A question was raised among scientists about exploring a technique that would not only be an efficient feature extractor but also hold the computational requirements under control.

State-space model (SSM) is an important concept that previously appeared mainly in control theory. Researchers' efforts in creating S4 [13] have made initial strides in applying SSMs to deep learning, mitigating issues compared to directly using conventional SSMs such as vanishing/exploding gradients and reduced efficiency with longer inputs. But focusing on the performance criterion, S4 still has a long way to go to catch up with Transformers. Thanks to Gu and Dao's introduction of State Space Models with Selective Scan (S6), also known as Mamba [14], the technique enhanced S4 with a selective mechanism that enables the model to emphasize pertinent data in an input-dependent approach. This method forms the crucial foundation for applying SSM to image tasks. Pioneering publications were Vision Mamba [15] and Visual Mamba [16], which focused on image classification then numerous studies have exhibited, showing potential to be the state-of the-art component of computer vision major. Even though there are abundant applications of Mamba in medical image segmentation such as UMamba [17], LightM-Unet [18], Weak-Mamba-unet [19], MambaBTS [20]; skin lesion segmentation tasks have remained a challenge for Mamba to deal with. Truthfulness is the most important criterion while working with skin lesion images; therefore, in this study, we introduce AC-MambaSeg, leveraging the strengths of CNN as well as Vision Mamba, to ultimately enable the model to capture both local and global information. Considering other techniques to enhance flexibility and adaptation of the proposed model, we use Convolutional Block Attention Module (CBAM) and Selective Kernels. The contributions of this work are unfolded as follows:

- Introducing a novel architecture, AC-MambaSeg, for skin lesion segmentation, which combines the strengths of VSS-CNN, CBAM, Attention gate, and Selective Kernels.
- Demonstrating significant improvements in segmentation performance compared to existing methods, as validated on ISIC2018 and PH2 benchmark datasets.

The remainder of this work is organised in the following manner. Section 2 presents the methodologies of our segmentation model. In the subsequent Section 3, the experimental findings are shown to validate the efficiency of the proposed network. The work is finally concluded in Sect. 4, which summarises the contributions and discusses possible directions for further research.

## 2. RELATED WORK

Visual State Space Model. Mamba [14] can first adapt and bring the State Space Model (SSM) method from control theory to Natural Language Processing field. In January 2024, Liu et al. published the first version of a pioneering publication [16] to adopt the state space model for a visual perception task. Visual Mamba defined a great new way to capture global information from an image and still managed the computational resource and time-consuming well. Due to its potential, the recent trend of applying SSM has gained significant momentum in visual-related tasks: classification [21], [22], video [23], event camera [24], [25], etc. In particular, in medical image segmentation, the application of SSM has led to notable advancements such as U-Mamba [17] constructed a hybrid CNN-SSM block combines the local feature extraction of convolutional layers with SSMs' ability to capture long-range dependencies, Lightm-Unet [18], Ultralight VM-Unet [26] introduced light models still have acceptable performance.

Convolutional Block Attention Module (CBAM). Convolutional Block Attention Module (CBAM) has emerged as a powerful mechanism for enhancing feature representations in convolutional neural networks (CNNs) [27]. CBAM leverages both channel-wise and spatial attention mechanisms to selectively emphasize informative features while suppressing irrelevant ones. By adaptively recalibrating feature maps, CBAM improves the discriminative power and generalization performance of CNNs, making it effective for tasks such as AVNC [28] for image classification, object detection with MTCNet [29], and segmentation with Cross-CBAM [30]. Especially in biomedical segmentation, CBAM holds a critical role when it has been applied in various applications: CBAM-Unet++ [31], CPA-Unet [32], and Rdau-net [33].

Selective Kernels (SK). In recent years, Selective Kernel (SK) units have gained attention in the field of computer vision for their ability to capture multi-scale contextual information [34]. These units dynamically adjust the receptive field size based on the task requirements, allowing models to selectively integrate features from different spatial regions. By adaptively modulating the convolutional properties, SK units enhance the model's capability to capture both local details and global context, leading to improved performance in segmentation tasks [35]–[37].

# 3. PROPOSED NETWORK ARCHITECTURE

## 3.1. Architecture Overview

In this section, we introduce the proposed AC-MambaSeg architecture, depicted in Fig. 1. Inheriting the legendary architecture of the segmentation task, which is U-Net [38], the model is built with a symmetric U-shaped configuration, comprising an encoder and a decoder connected by a bottleneck block and numerous advanced skip connections. Considering the main part of the model, Mamba and CNN complement each other perfectly to create an efficient hierarchical feature extractor at both locally and globally. A biomedical skin image $I \in \mathbb{R}^{C \times H \times W}$ will flow through six stages of feature extraction listed the input convolutional processing (*ConvIn*) and five consecutive encoder blocks. After each stage, the number of channels in the feature maps is doubled, accompanied by a halving of the resolution. As a result, expressions of deep feature maps of an input will be described as follows:

$$F_i \in \mathbb{R}^{C_i \times H_i \times W_i}, \ i \in [0, 5]$$

$$C_{i+1} = 2 \cdot C_i = 16 \cdot 2^{i+1}, \ H_{i+1} = \frac{H_i}{2}, \ W_{i+1} = \frac{W_i}{2} \tag{1}$$

in which, except $F_0$ is the feature map after the *ConvIn* processing stage, $F_i$ represents the output of encoder block $i^{th}$. Before being transferred to the next section, the deepest feature map is going underneath the *Bottleneck* block, while the skip feature at each encoded stage will be treated by CBAM component. Then the responsibilities of decoder blocks include combining and doing further feature extraction with features received from previous processes. After all, *ConvOut* creates the final segmentation mask corresponding to the input image.

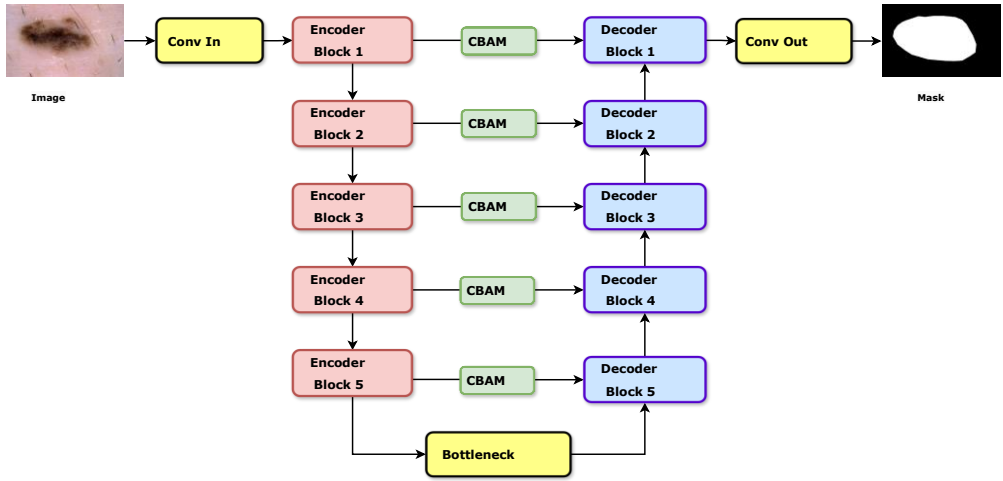

Fig. 1. Overall architecture of the proposed AC-MambaSeg network.

## 3.2. ResVSS Block

Leveraging breakthroughs of Visual Mamba, we proposed the ResVSS block in Fig. 2, as a main component of our architecture. Compared to the original Vanilla VSS Block, we have streamlined by removing an unnecessary Linear layer in the inner shortcut connections to reduce parameters and emphasize the capability of the core State Space 2D module. An additional previous depthwise convolution block gains more efficiency for the VSS block's performance, and a scaled residual connection, which is obtained after multiplying the original feature map with a learnable scale parameter, helps the information flow be coherent before and after the block. Instance Normalization (IN) [39] and ReLU are chosen to gain the model's nonlinearity and stability.

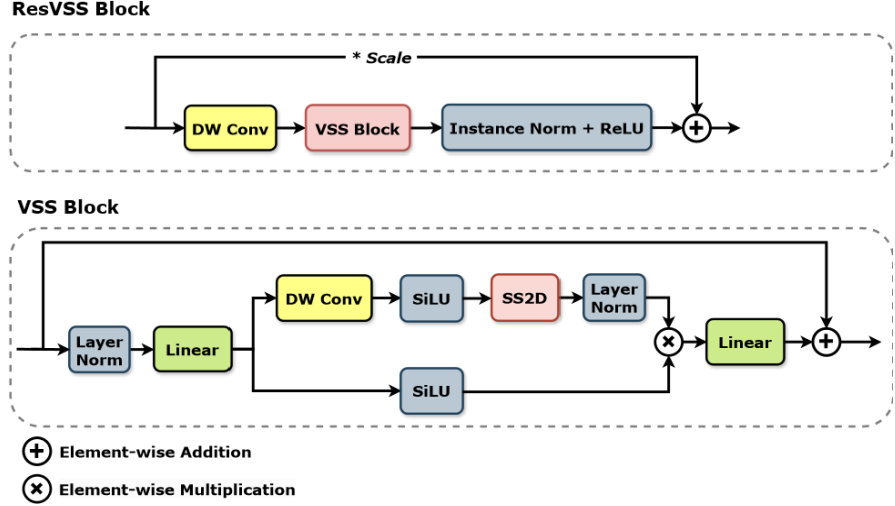

Fig. 2. Structure of ResVSS block

### 3.3. Encoder and Decoder Blocks

Association is the major intuition for our architecture. Both the encoder and decoder blocks are built based on a fusion of CNN and ResVSS, exploiting the CNN's robust local feature extraction capability and the VSS's extensive receptive field.

Considering the Encoder block shown in Fig. 3, input first undergoes a ResVSS block, then a normal 3x3 convolution layer, Batch Normalization layer, and a ReLU activation function. The feature map, before passing through MaxPooling to collapse the resolution, will be simultaneously used for the skip connection.

**Encoder Block**

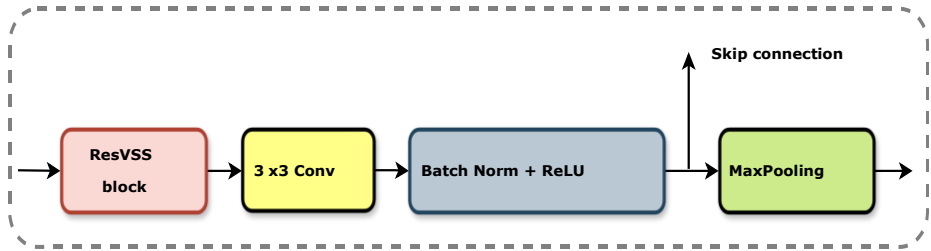

Fig. 3. The architecture of Encoder Block

For the Decoder half, each Decoder block receive features from Bottleneck or the preceding Decoder block, and from skip path of the opposite Encoder; process them with basic UpSampling method and Attention Gate [40]. An Attention Gate does not require heavy computational cost but enhance the skip connection quality pretty much. Aggregated feature traversed a 3x3 convolution layer, a Batch Normalization layer, and a ReLU function. At this point, the features have been significantly reduced in channel dimension before passing through ResVSS to avoid the potential surge in computational resources. All the above descriptions are visualized in Fig. 4.

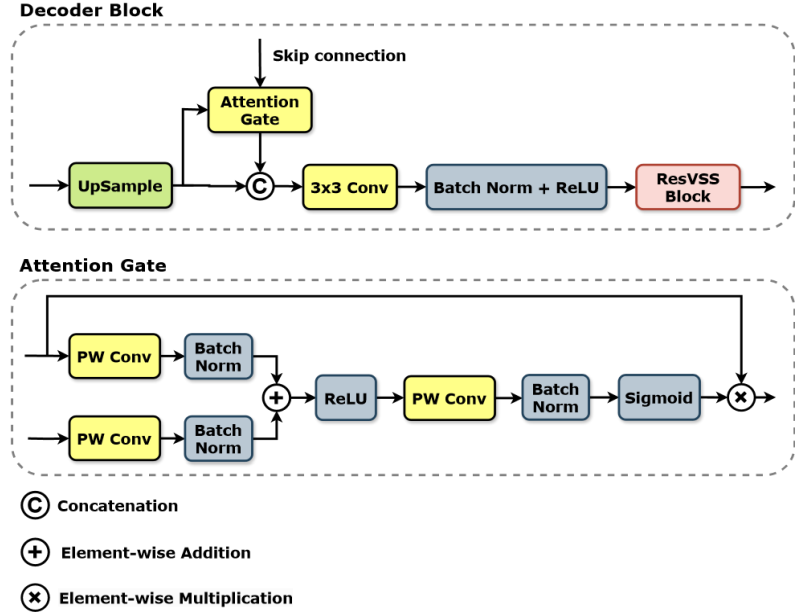

Fig. 4. The architecture of Decoder Block and Attention Gate

### 3.4. Skip Connection and Bottleneck

Skip connections and Bottleneck block are also essential parts of the model, bridge the information between encoder and decoder. Besides being refined by Attention Gate in Decoder, skip feature quality is enriched by Convolutional Block Attention Module (CBAM).

Fig. 5 represents the structure of the Bottleneck block. We are using Selective Kernels Convolution, in which the strategy dynamically adjusts the dilation of the 3x3 convolution based on the characteristics of each region. The Bottleneck contains two pointwise (PW) convolutions and a residual path to enable selective feature fusion and capturing any important information that may have been lost during the convolutional operations.

### 4. EXPERIMENT

### 4.1. Datasets

To assess the efficacy of the proposed AC-MambaSeg model, we utilize two well-known datasets in this domain, namely ISIC-2018 [41] from The International Skin Imaging Collaboration and PH2 [42] acquired at the Dermatology Service of Hospital Pedro Hispano.

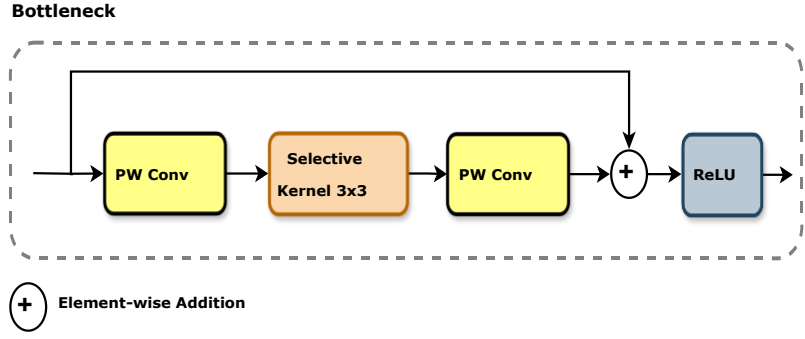

Fig. 5. The architecture of Bottlebeck Block

The ISIC-2018 dataset has a total of 2594 images, including dermatoscopic images accompanied by segmentation mask annotations. This dataset was subsequently partitioned into two subsets: 2074 images for training the model, and 520 images for model testing. Initially, the images had dimensions of 2016×3024 pixels, which were then resized to 192 × 256 pixels before being fed into the model.

The PH2 dataset has a total of only 200 images, including dermatoscopic images accompanied by segmentation mask annotations. This dataset was subsequently partitioned into two subsets: 170 images for training the model, and 30 images for model testing. Initially, the images had dimensions of 768 × 560 pixels, which were then resized to 192 × 256 pixels before feeding to the model.

**4.2. Training**

In this study, Adam optimization strategy is chosen for the training process, with an initial learning rate $lr = 2e-4$. The learning rate is halved if the model reaches a plateau in the Dice score after 10 epochs. Each dataset is trained from scratch for 200 epochs with a batch size of 8. The loss function is the combination between Dice loss and Tversky loss [43]. The mathematical expression for the Dice-Tversky Loss is given as follows:

$$\mathcal{L}(y,p) = 0.5 \cdot \mathcal{L}_{Dice} + 0.5 \cdot \mathcal{L}_{TV} \quad (2)$$

$$\mathcal{L}_{Dice}(y,p) = 1 - \frac{2\sum_{i=1}^{N} y_i p_i}{\sum_{i=1}^{N}(y_i + p_i)} \quad (3)$$

$$\mathcal{L}_{TV}(y,p) = 1 - \frac{2\sum_{i=1}^{N} y_i p_i}{\sum_{i=1}^{N} y_i p_i + \alpha \sum_{i=1}^{N} y_i \bar{p}_i + \beta \sum_{i=1}^{N} \bar{y}_i p_i} \quad (4)$$

where $y_i \in {0,1}$ and $p_i \in (0,1)$ denote the ground-truth value and predicted probability, respectively, for each pixel $i \in {1, 2,...,N}$ with $N$ is the total pixels number of the output segmentation mask.

**4.3. Evaluation Metrics**

To assess the model's performance, we employ the two predominant metrics in semantic segmentation: the Dice Similarity Coefficient (DSC) and Intersection over Union (IoU). Both metrics serve to gauge the resemblance between the ground truth and the predicted mask. The mathematical representations of DSC and IoU are outlined below:

$$DSC = \frac{2 \cdot TP}{2 \cdot TP + FP + FN + \varepsilon} \quad (5)$$

$$IoU = \frac{TP}{TP + FP + FN + \varepsilon} \quad (6)$$

with the terms *TP, FP,* and *FN* represent True Positives, False Positives, and False Negatives, respectively. The smooth coefficient $\varepsilon$ prevents zero division.

**4.4. Results and Comparison**

To evaluate the performance of the proposed approach, we conduct experiments on the ISIC18 and PH2 datasets and compare the results with other state of the arts. The compared models include Unet [38], SegNet [44], Attention Unet [40]: the U-shaped segmentation model that first introduced Attention Gate, Swin-Unet [10]: an Unet-like Pure Transformer method; and VM-Unet [45]: a model containing fully Vision Mamba block. Our suggested approach's qualitative visualisations for ISIC18 and PH2 are respectively displayed in Fig. 6 and Fig. 7. Our suggested method has correctly visualised a few of the assessment items, both easy and tough.

To evaluate the effectiveness of our proposed model, we compare its performance using the Dice Similarity Coefficient and Intersection over Union metrics with other methods. Additionally, to evaluate the effectiveness of the key components in the model, we compare the full model with models where adaptive components like CBAM and AG are removed or where the VSS block is replaced with a regular CNN layer. All these models were implemented based on the authors' publicly available code. Each model is trained under identical conditions as detailed in Section IV-B to ensure a fair and consistent comparison.

Initially, we evaluate our method on the ISIC2018 dataset. As depicted in Table I, we present the outcomes of our trained model under the specified conditions. Our proposed approach achieves impressive results, achieving a DSC of 0.9068 and an IoU of 0.8417. Comparing with traditional models mainly built from CNNs such as Unet, SegNet, or even Attention Unet, the proposed model demonstrates a superior performance improvement of up to approximately 5% in DSC. Swin-unet and VM-Unet have been applied by powerful techniques, namely Transformer and Mamba, so they performed quite well but still stay behind our AC-MambaSeg model.

TABLE I

COMPARISON OF THE PERFORMANCE AND COMPUTATIONAL EFFICIENCY OF THE PROPOSED MODEL AND OTHER MODELS ON THE ISIC 2018 DATASET.

| Model | Params | FLOPS | DSC | IoU |
|---|---|---|---|---|
| U-Net [38] | 31.2M | 36.2G | 0.8485 | 0.7633 |
| SegNet [44] | 29.4M | 30.1G | 0.8703 | 0.7894 |
| Attention Unet [40] | 34.9M | 49.9G | 0.8676 | 0.7911 |
| Swin-Unet [10] | 27.2M | 6.03G | 0.8792 | 0.8086 |
| VM-Unet [45] | 44.3M | 7.36G | 0.8935 | 0.8258 |
| Proposed | 8.0M | 2.09G | **0.9068** | **0.8417** |

Beyond its exceptional performance, our model stands out for its efficiency, requiring only 8.00 million parameters and a computational cost of 2.09 GFLOPS, significantly fewer than the approximately 20 million parameters required by the smallest of compared models.

Table II compares the performance of the proposed model with different configurations. The first row indicates the results without using CBAM and AG, as well as replacing VSS by a CNN layer, achieving the lowest statistics with a DSC of 0.8996 and an IoU of 0.8336. Remarkably, the third row showcases the performance without CBAM and AG but with VSS, yielding a higher DSC of 0.9027 and IoU of 0.8384, which demonstrate the strength of VSS block. These versions all have slightly fewer parameters and FLOPS, but we still opt for the combination of both VSS and CBAM, AG for the purpose of achieving the highest accuracy.

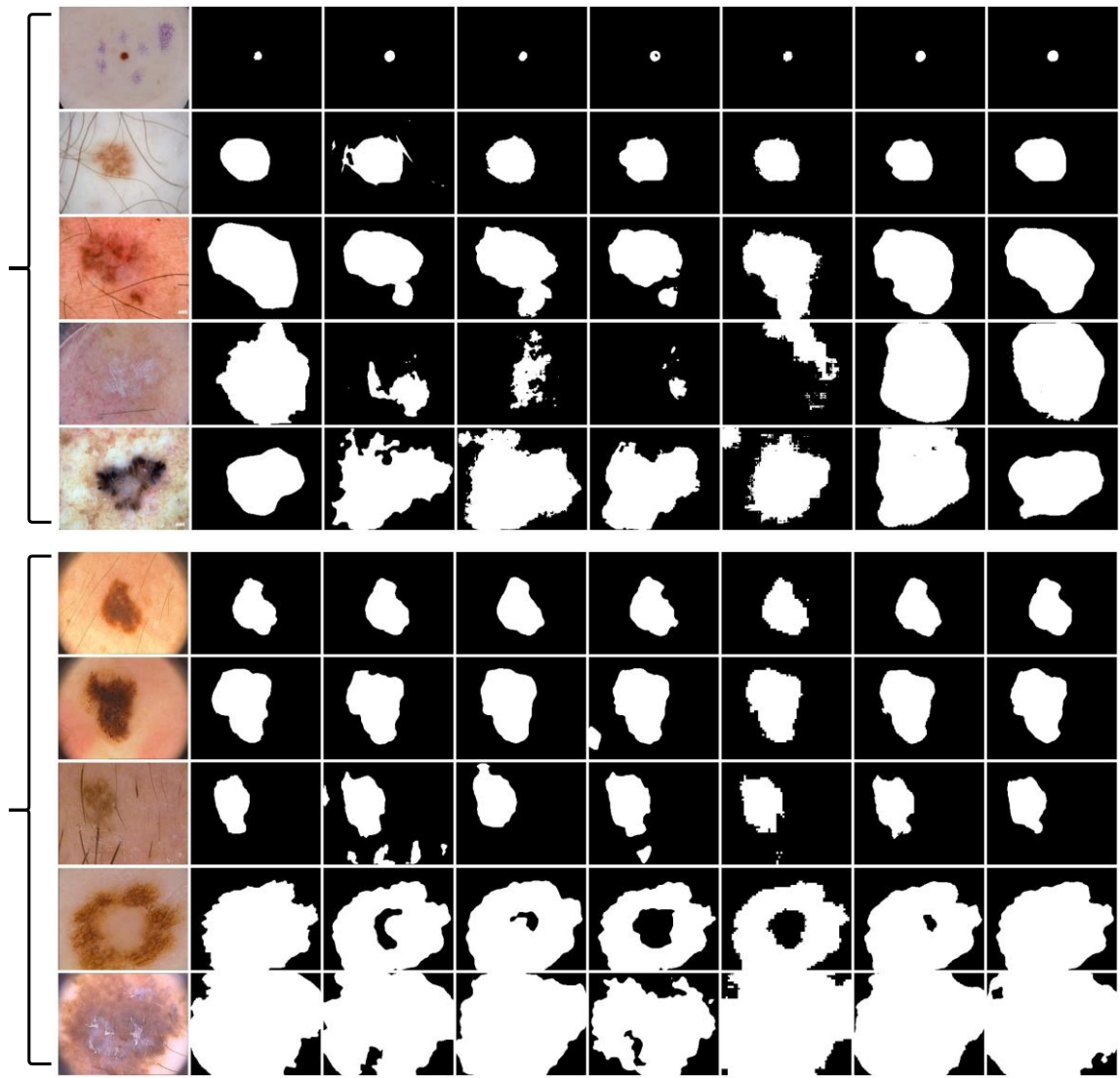

Fig. 6. Representative outcomes on ISIC2018 and PH2 datasets

TABLE II
COMPARISON OF THE PERFORMANCE AND COMPUTATIONAL EFFICIENCY TO
DEMONSTRATE THE EFFECTIVENESS OF THE
COMPONENTS WITHIN THE MODEL, ON THE ISIC 2018 DATASET.

| Model | Params | FLOPS | DSC | IoU |
| --- | --- | --- | --- | --- |
| w/o CBAM, AG; w/o VSS | 7.52M | 0.06G | 0.8784 | 0.8114 |
| w CBAM, AG; w/o VSS | 7.92M | 0.33G | 0.8996 | 0.8323 |
| w/o CBAM, AG; w VSS | 7.60M | 1.82G | 0.9027 | 0.8384 |
| Proposed | 8.00M | 2.09G | **0.9068** | **0.8417** |

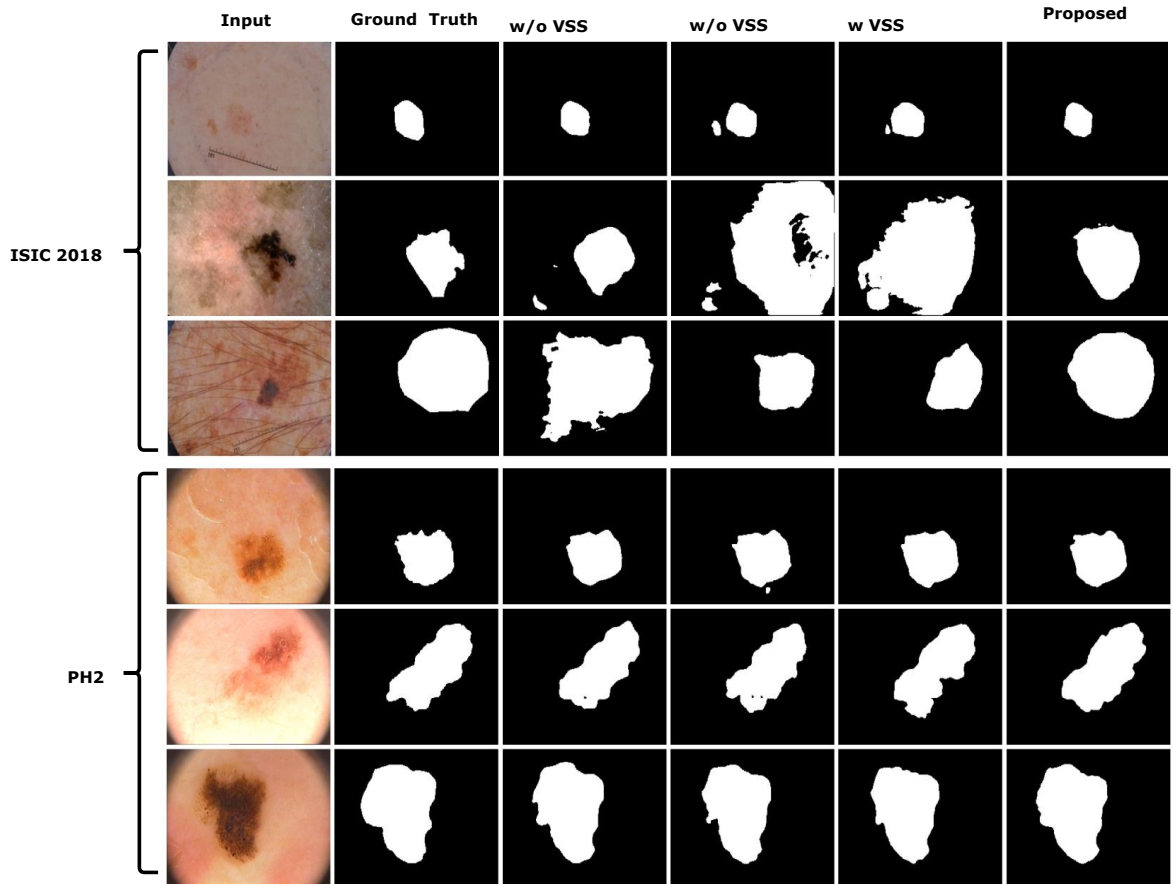

Fig. 7. Representative outcomes of proposed model in different configurations

Similar to the ISIC2018 dataset, Tables III and IV present a comparison of model performance and computational efficiency on the PH2 dataset. The first table showcases the DSC and IoU scores along with the number of parameters and FLOPS for various models, including U-Net, SegNet, Attention Unet, Swin-Unet, VM-Unet, and the proposed model. Notably, the proposed model outperforms others with the highest DSC and IoU scores which are 0.9544 and 0.8417, respectively, while maintaining a relatively low number of parameters and FLOPS.

TABLE III
COMPARISON OF THE PERFORMANCE AND COMPUTATIONAL
EFFICIENCY OF THE PROPOSED MODEL AND OTHER MODELS ON THE PH2 DATASET.

| Model | Params | FLOPS | DSC | IoU |
| --- | --- | --- | --- | --- |
| U-Net [38] | 31.2M | 36.2G | 0.8994 | 0.8971 |
| SegNet [44] | 29.4M | 30.1G | 0.9130 | 0.8735 |
| Attention Unet [40] | 34.9M | 49.9G | 0.9101 | 0.8803 |
| Swin-Unet [10] | 27.2M | 6.03G | 0.9337 | 0.8779 |
| VM-Unet [45] | 44.3M | 7.36G | 0.9298 | 0.8969 |
| Proposed | 8.0M | 2.09G | **0.9544** | **0.9146** |

TABLE IV
COMPARISON OF THE PERFORMANCE AND COMPUTATIONAL EFFICIENCY TO
DEMONSTRATE THE EFFECTIVENESS OF THE COMPONENTS WITHIN THE MODEL, ON THE
PH2 DATASET.

| Model | Params | FLOPS | DSC | IoU |
| --- | --- | --- | --- | --- |
| w/o CBAM, AG; w/o VSS | 7.52M | 0.06G | 0.9124 | 0.8928 |
| w CBAM, AG; w/o VSS | 7.92M | 0.33G | 0.9187 | 0.8798 |
| w/o CBAM, AG; w VSS | 7.60M | 1.82G | 0.9426 | 0.9050 |
| Proposed | 8.00M | 2.09G | **0.9544** | **0.9146** |

In Table IV, different configurations of the proposed model are evaluated by adjusting the presence of CBAM, AG, and VSS components. Each configuration is assessed based on DSC, IoU, parameters, and FLOPS. Despite variations in component inclusion, the proposed model consistently achieves superior performance compared to other configurations, reaffirming the effectiveness of its design choices.

## 5. CONCLUSION

This study highlights the efficacy of the proposed ACMambaSeg, leveraging the robustness of VSS-CNN and the adaptability of Mamba to enhance skin lesion segmentation. The incorporation of CBAM, Attention Gate, and Selective Kernels further strengthens the model's performance, as evidenced by significant improvements over existing methods on benchmark datasets like ISIC2018 and PH2. Future research efforts will explore the full potential of AC-MambaSeg, particularly in optimizing Mamba and VSS integration to unlock even greater segmentation accuracy and versatility in dermatological imaging applications.